# Principle Based Semantics for HPSG

**Anette Frank** and **Uwe Reyle**
Institute for Computational Linguistics
University of Stuttgart
Azenbergstr.12, D–70174 Stuttgart, Germany
e-mail: uwe@ims.uni-stuttgart.de

## Abstract

The paper presents a constraint based semantic formalism for HPSG. The syntax-semantics interface directly implements syntactic conditions on quantifier scoping and distributivity.[1] The construction of semantic representations is guided by general principles governing the interaction between syntax and semantics. Each of these principles acts as a constraint to narrow down the set of possible interpretations of a sentence. Meanings of ambiguous sentences are represented by single partial representations (so-called U(nderspecified) D(iscourse) R(epresentation) S(tructure)s) to which further constraints can be added monotonically to gain more information about the content of a sentence. There is no need to build up a large number of alternative representations of the sentence which are then filtered by subsequent discourse and world knowledge. The advantage of UDRSs is not only that they allow for monotonic incremental interpretation but also that they are equipped with truth conditions and a proof theory that allows for inferences to be drawn directly on structures where quantifier scope is not resolved.

## 1 Introduction

The semantic analysis of standard HPSG deviates from the familiar Montegovian way to construct semantic representations mainly in that it uses unification to eliminate the need for $\beta$-reduction. Variables are bound to argument positions by the close interplay between syntactic and semantic processing; and the semantics of constituents is determined by the Semantics Principle, which governs the way of unifying the semantics of daughter constituents to build up the semantic value of the phrasal constituent: The CONTENT value is projected from the *semantic head*, which is defined as the syntactic HEAD-DTR in head-comp-structures, but as the ADJ-DTR in head-adjunct structures. It is important to note that the semantic contribution of quantified verb arguments is not completely projected as part of the CONTENT value. The meaning of such NPs splits into the features QUANTS, a list representing the information about quantifier scope, and NUCLEUS, containing the nonquantificational core. In the general case only the NUCLEUS is projected from the semantic head according to the Semantics Principle, while the QUANTS value gets instantiated stepwise in interaction with the quantifier storage mechanism (Cooper Store). The mechanism of Cooper storage is built into HPSG by use of two further attributes, QSTORE and RETRIEVED, both represented as sets of quantifiers. All quantifiers start out in QSTORE by lexical definition. The Semantics Principle defines the inheritance of QSTORE to the phrasal constituents, where they may be taken out of store by an appropriately instantiated RETRIEVED value and then put into the QUANTS value of the CONTENT feature. The order in which the semantic value of quantified NPs is retrieved fixes their relative scope. To analyse sentences with scope ambiguities several parses are thus necessary. Besides the definition of appropriate restrictions to and configurations for applications of RETRIEVED the main problem we face with this kind of analysis is to modify the semantics of HPSG in such a way that it yields underspecified representations and not sets of fully specified ones. Further shortcomings of HPSG semantics are the following. First, adjuncts (like quantificational adverbs, modals) and also negation bear the potential to introduce scope ambiguities. In order to treat them by the same mechanism that treats the arguments of the verb their meaning representation would ha-ve to be put into store. This, however, requires further modifications of the Semantics Principle, because the treatment of head-adjunct structures differs essentially from the treatment of other configurations (see (Pollard/Sag 1994), Ch.8).[2] Second, there is no underspecified representation of ambiguities that arise from the distributive/collective distinction of plural NPs (neither within the HPSG framework nor in the C(ore)L(anguage)E(ngine)[3]). Third, the semantic representation of indefinite NPs must be independent of the context in which they are interpreted. We do not want to switch from a universally quantified interpretation to an existentially quantified one, when we come to disambiguate the ambiguous sentence **Every student who admires a philosopher reads his original writings** such that **a philosopher** is interpreted specifically. This requirement calls for DRT as underlying semantic formalism.

---

[1] In the present paper we do only focus on simple principles restricting scope ambiguities and ambiguities resulting from plural NPs in English. For German restrictions on scope are much more complicated because they cannot be stated independently of scrambling phenomena. In (Frank/Reyle 1994) the present approach is worked out for a fragment of German that deals with (i) quantifier scope ambiguities triggered by scrambling and/or movement and (ii) ambiguities that arise from the collective/distributive distinction of plural NPs. The underlying scope theory for German was developed in (Frey 1993). The analysis in (Frank/Reyle 1994) departs significantly from our earlier account in (Frank/Reyle 1992), where monotonicity was not ensured.

In the sequel of this paper we show how the extension of DRT to UDRT developed in (Reyle 1993) can be combined with an HPSG-style grammar. The basic idea of the combination being that syntax as well as semantics provide structures of equal right; that the principles internal to the syntactic and semantic level are motivated *only* by the syntactic and semantic theory, respectively; and that mutually constraining relations between syntax and semantics are governed by a separate set of principles that relate syntactic and semantic information appropriately. We will replace the Semantics Principle of standard HPSG versions by a principle which directly reflects the *monotonicity* underlying the interpretation process designed in (Reyle 1993): At any stage of the derivation more details are added to the description of the semantic relations between the various components of the sentence, i.e. the partial representation of any mother node is the union of the partial representations of its daughter nodes plus further constraints derived from the syntactic, semantic and also pragmatic context.

## 2 Quantifier Scope and Partial Orders

The need for underspecified representations is by now widely accepted within computational and theoretical linguistics.[4] To make the results of the ongoing research on underspecified representations available for HPSG we may pursue two strategies. According to the first strategy we take the HPSG-style analysis – essentially as it is – and only apply slight modifications to produce underspecified output. The second strategy involves a more radical change as it takes an existing theory of underspecified representations and replaces the HPSG semantics by the construction principles of this theory.

Let us start out with a sketch of the first approach. It will show us where its limitations are and allow us to compare different approaches to underspecification. The first thing to do, when un-specifying HPSG semantics, is to relax the retrieval operation. This must be done in two respects. First, we must allow NP-meanings not to be retrieved at all. This results in their relative scope not being determined. Second, we must accommodate syntactic and semantic restrictions on possible scope relations to be stated by the grammar.[5] Restrictions specifying, for example, that the subject NP must always have wide scope over the other arguments of the verb; or, that the scope of genuinely quantified NPs is clause bounded. The modifications we propose are the following. First, we incorporate the QSTORE feature into the CONTENT feature structure. This makes the NP meanings available even if they are not retrieved from QSTORE. Second, we take the value of the QUANTS feature not to be a "stack" (i.e. by appending new retrieved quantifiers as first elements to QUANTS), but allow any NP meaning that is retrieved at a later stage to be inserted *at any place* in that list. This means that the order of NP meanings in QUANTS fixes the relative scope of these meanings only; it does not imply that they have narrow scope with respect to the NP meaning that will be retrieved next. But this is not yet enough to implement clause boundedness. The easiest way to formulate this restriction is to prohibit projection of quantified NP meanings across bounding nodes. Thus the QSTORE and QUANTS values of a bounding node inherit the quantificational information only of *indefinite* NPs and not of *generalized quantifiers*. To be more precise, let us consider the tree $\beta$ consisting only of the bounding nodes in the syntactic analysis of a sentence $\gamma$. Then the semantic content of $\gamma$ can be associated with nodes of $\beta$ in the following way. For each node $i$ of $\beta$ the attributes QUANTS, QSTORE and NUCLEUS have values $quants_i$, $qstore_i$ and $nucleus_i$. The relative scope between scope bearing phrases of $\gamma$, i.e. between the elements of $\bigcup_i (quants_i \cup qstore_i)$ can then be defined as follows.

- If $Q_1$ and $Q_2$ are in $quants_i$ and $Q_1$ precedes $Q_2$, then $Q_1$ has scope over $Q_2$.
- If $Q_1$ is in $quants_i$ and $Q_2$ in $quants_j$, where $i$ dominates $j$, then $Q_1$ has scope over $Q_2$.
- If $Q_1$ is in $qstore_i$ and not in $qstore_j$, where $i$ dominates $j$, then $Q_1$ has scope over any $Q_2$ in $qstore_j \cup quants_j$ that are not in $qstore_i \cup quants_i$.

---

[2]For general criticism of the analysis of adjuncts in standard HPSG see (Abb/Maienborn 1994). Their analysis of adjuncts in HPSG fits neatly into the account of semantics projection to be presented below.

[3]See (Alshawi 1992). In CLE the resolution of QLFs also involves disambiguation with respect to this kind of ambiguities.

[4]See (Peters/vanDeemter 1995) for recent discussion.

[5]This has to be done also for the standard theory.

The last clause says that any NP $Q_1$ occurring in the clause of level $i$ and that is still in QSTORE has scope over all quantified NPs $Q_2$ occurring in embedded clauses (i.e. clauses of level $j$). But $Q_1$ does not necessarily have scope over any indefinite NP introduced at level $j$.

Those familiar with the work of Alshawi and Crouch (Alshawi/Crouch 1992) might have noticed the similarity of their interpretation mechanism and what we have achieved by our modifications to standard HPSG semantics. The elements of QUANTS play exactly the same role as the instantiated metavariables of Alshawi and Crouch. This means that we could adapt their interpretation mechanism to our partially scoped CONTENT structures. But note that we already have achieved more than they have as we are able to express the clause-boundedness restriction for generalized quantifiers.

We will not go into the details and show how the truth conditions of Alshawi and Crouch have to be modified in order to apply to partially scoped CONTENT structures. We will instead go ahead and work out the limitations of what we called the first strategy. To keep things as easy as possible we restrict ourselves to the case of simple sentences (i.e. to trivial tree structures of QSTORE and QUANTS values that consist of one single node only). In this case the QUANTS value (as well as the instantiation of metavariables) imposes a partial order on the relative scope of quantifiers. Assume we had a sentence with three quantifiers, $Q_1$, $Q_2$ and $Q_3$. Then the possible lenghts of QUANTS values varies from 0 to 3. Lengths 0 and 1 leave the relative scope of $Q_1$, $Q_2$ and $Q_3$ completely underspecified. Values of length 2 say that their first element always has wide scope over the second, leaving all possible choices for the third quantifier. And finally we have the fully specified scoping relations given by values of length 3. There are, however, some possibilities to restrict scope relationships that cannot be represented this way: One cannot, for example, represent the ambiguity that remains if we (or, syntax and semantics) require that $Q_1$ and $Q_2$ must have scope over $Q_3$, but leaves unspecified the relative scope between $Q_1$ and $Q_2$; nor are we able to express a restriction that says $Q_1$ must have scope over both, $Q_2$ and $Q_3$, while leaving the relative scope between $Q_2$ and $Q_3$ unspecified. Retrieving a quantifier $Q_i$ (or starting to calculate the truth value of a sentence by first considering this quantifier) is an operation that takes $Q_i$ and adds it to QUANTS. As QUANTS is a list this amounts to a full specification of the relative scope of $Q_i$ with respect to *all* other elements already contained in QUANTS. This shows that the expressive power of the representation language is too restrictive already for simple sentences. We need to represent *partial* orders of quantifier scope. But we cannot do this by talking about a pair consisting of a quantifier $Q_i$ and a list of quantifiers QUANTS. We must be able to talk about *pairs of quantifiers*. This not only increases the expressive power of the representation language, it also allows for the formulation of restrictions on quantifier scope in a declarative *and* natural way. The formalism of UDRSs we introduce in the following section is particularly suited to 'talk' about semantic information contributed by different components of a sentence. It therefore provides a particularly good ground to implement a principle based construction of semantic representations.

## 3   UDRS Construction in HPSG

In the following we will design a syntax-semantics interface for the construction of UDRSes in HPSG, focussing on the underspecified representation of scope and plural. To overcome the problems discussed in Section 2 we chose to depart from the semantics used in standard HPSG (Pollard/Sag 1994), and instead allow for the construction of (U)DRSes. The structure of the CONTENT attribute as well as the Semantics Principle will be changed substantially, since the construction of (U)DRSes allows for inherently different information structures and processing mechanisms. The former CONTENT attribute is replaced by a complex feature structure UDRS, consisting of three attributes, LS, SUBORD and CONDS.

(1) $\left[ \text{LOC} \begin{bmatrix} \text{CAT } cat \\ \text{UDRS} \begin{bmatrix} \text{LS} \begin{bmatrix} \text{L-MAX } \mathbf{l}_{max} \\ \text{L-MIN } \mathbf{l}_{min} \end{bmatrix} \\ \text{SUBORD } \{\mathbf{l} \leq \mathbf{l'}, ...\} \\ \text{CONDS } \{\gamma_i, ...\} \end{bmatrix} \end{bmatrix} \right]$

CONDS is a set of labelled DRS-conditions, $\gamma_i$, the form of which is determined by lexical entries. SUBORD contains information about the hierarchical structure of a DRS. It is expressed by means of a subordination relation, $\leq$, between labels. If $\gamma_1$ and $\gamma_2$ are two DRS-conditions with labels $l_1$ and $l_2$ such that $l_1 \leq l_2$ is contained in SUBORD, then this is equivalent to saying that $\gamma_1$ and $\gamma_2$ will occur in DRSs $K_1$ and $K_2$ such that $K_1$ is weakly subordinate to $K_2$, i.e. $K_1$ is either identical to $K_2$ or nested within it. SUBORD thus imposes the structure of an upper semi-lattice with one-element, $l_\top$, to the set of labels. The attribute LS defines the distinguished labels, which indicate the upper and lower bounds for a DRS-condition within the semilattice.

The main task in constructing UDRSes consists in appropriately relating the labels of the DRS-conditions that are to be combined. This is performed by the association of DRS-conditions with distinguished labels in the lexical entries on the one hand and by conditions governing the projection of the distinguished labels on the other. The role of the distinguished labels is most transparent with verbs and quantifiers.

In the lexical entry of a transitive verb, for example, the DRS-condition stated in CONDS is a relation holding between discourse referents.[6] This condition is associated with an identifying label **l**. In addition **l** is identified as the minimal distinguished label of the verbal projection by coindexation with L-MIN.

$$(2) \begin{bmatrix} \text{CAT} \mid \text{H} \mid \text{SC} < \begin{bmatrix} \text{CASE } nom \\ \text{DREF } \boxed{x} \end{bmatrix}, \begin{bmatrix} \text{CASE } acc \\ \text{DREF } \boxed{y} \end{bmatrix} > \\ \text{UDRS} \begin{bmatrix} \text{LS} \begin{bmatrix} \text{L-MIN } \boxed{1} \end{bmatrix} \\ \text{SUBORD } \{\} \\ \text{CONDS} \left\{ \begin{bmatrix} \text{LABEL } \boxed{1} \\ \text{REL } hire \\ \text{ARG1 } \boxed{x} \\ \text{ARG2 } \boxed{y} \end{bmatrix} \right\} \end{bmatrix} \end{bmatrix}$$

Generalized quantifiers, as in (3), introduce two new labels which identify the DRS-conditions of their restrictor and nuclear scope. The quantificational relation holding between them is stated in terms of the relation attribute, REL. In the lexical entry for *every*, given in (3), a new discourse referent is introduced in the restrictor DRS, labelled $l_{11}$, which is identified with the label of the subcategorized NP. The feature SUBORD defines the labels of restrictor and scope to be subordinate to the label $l_1$ which identifies the entire condition. The label $l_1$ is defined as the upper bound, or distinguished maximal label of the quantificational structure, whereas the lower bound, or distinguished minimal label is given by the label of the nuclear scope, $l_{12}$.

$$(3) \begin{bmatrix} \text{CAT} \begin{bmatrix} \text{HEAD } quant \\ \text{COMPS } < \text{NP} \begin{bmatrix} \text{LABEL } \boxed{l_{11}} \end{bmatrix} > \end{bmatrix} \\ \text{UDRS} \begin{bmatrix} \text{LS} \begin{bmatrix} \text{L-MAX } \boxed{l_1} \\ \text{L-MIN } \boxed{l_{12}} \end{bmatrix} \\ \text{SUBORD } \{ \boxed{l_1} > \boxed{l_{11}}, \boxed{l_1} > \boxed{l_{12}} \} \\ \text{CONDS} \left\{ \begin{bmatrix} \text{LABEL } \boxed{l_1} \\ \text{REL } every \\ \text{RES } \boxed{l_{11}} \\ \text{SCOPE } \boxed{l_{12}} \end{bmatrix}, \begin{bmatrix} \text{LABEL } \boxed{l_{11}} \\ \text{DREF } x \end{bmatrix} \right\} \end{bmatrix} \end{bmatrix}$$

The entry for the indefinite singular determiner, (4), introduces a new individual type referent. As indefinites do not introduce any hierarchical structure into a DRS the identity statement $l_1 = l_{12}$ for the minimal and maximal labels is defined in SUBORD.

$$(4) \begin{bmatrix} \text{CAT} \begin{bmatrix} \text{HEAD} \begin{bmatrix} \text{AGR} \mid \text{NUM } sg \end{bmatrix} \\ \text{COMPS } < \text{NP} \begin{bmatrix} \text{LABEL } \boxed{l_{12}} \end{bmatrix} > \end{bmatrix} \\ \text{UDRS} \begin{bmatrix} \text{LS} \begin{bmatrix} \text{L-MAX } \boxed{l_1} \\ \text{L-MIN } \boxed{l_{12}} \end{bmatrix} \\ \text{SUBORD } \{ \boxed{l_1} = \boxed{l_{12}} \} \\ \text{CONDS} \left\{ \begin{bmatrix} \text{LABEL } \boxed{l_1} \\ \text{DREF } x \end{bmatrix} \right\} \end{bmatrix} \end{bmatrix}$$

The construction of UDRSes will be defined in terms of clauses of the Semantics Principle: In (5), clause (I) of the Semantics Principle defines the inheritance of the partial DRSes defined in the CONDS attributes of the daughters to the CONDS value of the phrase. Contrary to the Semantics Principle of (Pollard/Sag 1994) the semantic conditions are always inherited from *both* daughters (we assume binary branching) and therefore project to the uppermost sentential level. Furthermore, clause (I) applies to *head-comp-* and *head-adj-structures* in exactly the same way.[7] Clause (II) of the Semantics Principle defines the inheritance of subordination restrictions: The subordination restrictions of the phrase are defined by the union of the SUBORD values of the daughters. Clause (III) of the Semantics Principle states the distinguished labels LS of the phrase to be identical to the distinguished labels of the HEAD-daughter. It is therefore guaranteed that in binary branching structures the minimal and maximal labels of the head category are available all along the (extended) head projection.[8] This prepares clauses (IV) and (V) of the Semantics Principle, which define the binding of discourse markers and locality of quantificational scope, respectively. We will first consider clause (IV) and will come back to clause (V) in the next Section.

In a (U)DRS, the partial structure of the verb has to be (weakly) subordinate to the scope of all the partial DRSes that introduce the discourse markers corresponding to the verb's arguments. This guarantees that all occurrences of discourse markers are properly bound by some superordinated DRS. The constraint is realized by clause (IV) of the Semantics Principle, the Closed Formula Principle. It guarantees that the label associated with the verb, which is identified with the distinguished minimal label of the sentential projection, is subordinated to the minimal label, or lower bound of each of the verb's arguments. Note that with quantified arguments the predicate of the verb must be subordinate to the nuclear scope of the quantifier. As defined in (3), it is in fact the nuclear scope of the quantified structure that will be accessed by the distinguished minimal label of the quantified NP. Thus the Closed Formula Principle (IV) in (5) states that in every (non-functional) *head-comp-struc* a further subordination restriction is unioned to the phrase's SUBORD value, which subordinates the minimal label of the head –here the minimal label associated with the verb– to the minimal label of its actual complement, which in case of a quantified argument identifies the nuclear scope.

**Semantics Principle:**[9]

$$(5) \begin{bmatrix} ..\text{UDRS} \begin{bmatrix} \text{LS } \boxed{5} \\ \text{SUBORD } ... \cup \{ \boxed{l_{min}} \geq \boxed{l_{verb}} \} \cup \boxed{3} \cup \boxed{4} \\ \text{CONDS } \boxed{1} \cup \boxed{2} \end{bmatrix} \end{bmatrix}_{head-comp-struc}$$

```
        C-DTR                      H-DTR
[..UDRS [LS [L-MIN l_min]      [..UDRS [LS 5 [L-MIN l_verb]
         SUBORD 4                       SUBORD 3
         CONDS 2 ]]                     CONDS 1 ]]
```

---

[6] The reference to discourse referents of the syntactic arguments is only provisionally stated here. For the precise definition see (10) below. The use of SUBCAT (SC) as a head attribute is motivated in (Frank 1994).

[7] See (Abb/Maienborn 1994) for a corresponding analysis of adjuncts.

[8] Functional categories inherit the distinguished labels of their complement (see (7)). The distinguished labels therefore project along the *extended* head projection.

I   Inheritance of UDRS-Conditions
II  Inheritance of subordination restrictions[10]
III Projection of the distinguished labels
IV  Closed Formula Principle

Note that generalized quantifiers were marked as *scope bearing* by non-identical values of minimal and maximal labels; and singular indefinite NPs were marked as *not scope bearing* by identifying minimal and maximal labels. As plural NPs introduce a quantificational condition when they are interpreted distributively but behave like indefinites when interpreted collectively, in a representation of their meaning that is underspecified with respect to the distributive/collective ambiguity plural NPs must be marked as *potentially scope bearing*. This can be achieved if in the lexicon entry of a plural determiner (6) we do not completely specify the relation between the minimal label $l_{12}$ and the maximal label $l_1$, but only require that $l_{12}$ is weakly subordinate to $l_1$. This weak subordination relation will be further restricted to either identity or strict subordination when more information is available from the semantic or pragmatic context that allows the ambiguity to be resolved. By monotonically adding further constraints a collective or quantificational (distributive or generic) reading of the plural NP may then be specified.[11] If a distributive reading is chosen, the minimal label $l_{12}$ will identify the nuclear scope of the quantified structure, and in the case of a collective reading the relation of (weak) subordination between minimal and maximal label will be reduced to identity. We will state this in detail in Section 4.

(6) $\begin{bmatrix} \text{CAT} & \begin{bmatrix} \text{HEAD} & [\text{AGR} \mid \text{NUM} \; pl] \\ \text{COMPS} & < [\text{LABEL} \; \boxed{l_1}] > \end{bmatrix} \\ \text{UDRS} & \begin{bmatrix} \text{LS} & \begin{bmatrix} \text{L-MAX} & \boxed{l_1} \\ \text{L-MIN} & \boxed{l_{12}} \end{bmatrix} \\ \text{SUBORD} & \{ \boxed{l_1} \geq \boxed{l_{12}} \} \\ \text{CONDS} & \{ \begin{bmatrix} \text{LABEL} & \boxed{l_1} \\ \text{DREF} & X \end{bmatrix} \} \end{bmatrix} \end{bmatrix}$

Together with the structure of the lexical entries illustrated above, the clauses (I) – (IV) of the Semantics Principle given in (5) define the core mechanism for UDRS construction: The Semantics Principle defines the inheritance of the labelled DRS conditions and of the subordination restrictions between these labels, which define the semilattice for the complete UDRS structure. The subordination restrictions are projected from the lexicon or get introduced monotonically, e.g. by the Closed Formula Principle to ensure the correct binding of discourse referents. Further subordination restrictions will be added – monotonically – by the remaining clauses of the Semantics Principle, to be introduced in the next Section.

## 4 Quantifier Scope and Plural Disambiguation

**Quantificational Scope** Since the conditions on quantificational scope for generalized quantifiers and distributive readings of plural NPs are dependent on syntactic structure, the Semantics Principle will be supplemented by further clauses governing the interface between syntactic constraints and semantic representation. Note that genuine quantifiers as well as distributive readings of plural NPs differ in their scope potential from indefinite NPs and collectively interpreted plural NPs. Whereas the latter may take arbitrarily wide scope, the scope of the former is clause bounded, i.e. they are allowed to take scope only over elements that appear in their local domain. We implement this restriction by requiring that the maximal label of a generalized quantifier be subordinate to the distinguished label that identifies the upper bound of the local domain. For plural NPs, a similar constraint must be stated in case a distributive reading is chosen which specifies the plural NP as *scope bearing*.

The distinction between *scope bearing* and *not scope bearing* NPs was defined by strict subordination and identity of the distinguished labels, respectively. In case a distributive reading is chosen by the clauses for plural disambiguation, to be stated below, the relation of weak subordination in (6), is strengthened to strict subordination. Yet, plural disambiguation may take place rather late in subsequent discourse, while the syntactic constraints for quantificational scope can only be determined locally. The Quantifier Scope Principle (V) will therefore introduce *conditionalized subordination restrictions* to define the clause-boundedness of both generalized quantifiers and distributively quantified plural NPs. [12]

For finite sentences the local domain for quantified verb arguments comes down to the local IP projection (Frey 1993). In a functional HPSG grammar (see (Frank 1994)) this local domain corresponds to the functional projection of the finite VP. The distinguished maximal label $l_{max}$ which identifies the upper bound of the local domain for quantified verb arguments will therefore be instantiated by the complementizer heading a finite sentence, as in (7).

(7) $\begin{bmatrix} \text{LOC} & \begin{bmatrix} \text{CAT} & \begin{bmatrix} \text{COMPS} < \begin{bmatrix} \text{VFORM} & fin \\ \text{LS} & \boxed{1} \end{bmatrix} > \end{bmatrix} \\ \text{UDRS} & \begin{bmatrix} \text{LS} & \boxed{1} & [\text{L-MAX} \; l_{max}] \end{bmatrix} \end{bmatrix} \\ \scriptstyle{func-cat} \end{bmatrix}$

---

[9] The Semantics Principle will only be given for *head-comp-structures*. For *head-subj-* and *head-adj-structures* corresponding clauses have to be stated. For *head-filler-structures* we only define inheritance of CONDS, SUBORD, and LS from the HEAD-DTR.

[10] The dots indicate that further subordination restrictions will be unioned to the phrase's SUBORD value by clause (V) of the Semantics Principle, defined below.

[11] We are not in the position to discuss the factors that determine these constraints here.

[12] The scoping principles described in (Frank/Reyle 1994) further account for the scope restrictions of generalized quantifiers and distributive plural NPs.

Due to the projection of the distinguished labels by clause (III) of the Semantics Principle and the definition of functional categories, the upper bound for the local domain of quantifier scope, $l_{max}$, is available throughout the extended projection, where clause (V) of the Semantics Principle, the Quantifier Scope Principle, applies. In (8), the Quantifier Scope Principle (V) states that if the complement is a generalized quantifier (type *quant*) or a *potentially scope bearing* plural NP (type *plural*) the SUBORD value of the phrase will contain a further *conditionalized* subordination constraint, which states that – if the argument is, or will be characterized as a *scope bearing* argument by strict subordination of its minimal and maximal label – the complement's maximal label $l_{quant}$ is subordinate to the label $l_{max}$ which identifies the upper bound of the local domain.

**Semantics Principle:**
Clauses I – IV & V Quantifier Scope Principle

$$(8) \quad \begin{bmatrix} \text{..UDRS} \begin{bmatrix} \text{LS } \boxed{5} \\ \text{SUBORD } \{\boxed{l_{quant}} > \boxed{l_{min}} \Rightarrow \boxed{l_{max}} \geq \boxed{l_{quant}}\} \\ \cup \{\boxed{l_{min}} \geq \boxed{l_{verb}}\} \cup \boxed{3} \cup \boxed{4} \\ \text{CONDS } \boxed{1} \cup \boxed{2} \end{bmatrix} \\ \text{head-comp-struc} \end{bmatrix}$$

C-DTR      H-DTR

$\begin{bmatrix} \text{CAT | HEAD } \text{quant} \lor \text{plural} \\ \text{UDRS} \begin{bmatrix} \text{LS } \begin{bmatrix} \text{L-MAX } \boxed{l_{quant}} \\ \text{L-MIN } \boxed{l_{min}} \end{bmatrix} \\ \text{SUBORD } \boxed{4} \\ \text{CONDS } \boxed{2} \end{bmatrix} \end{bmatrix}$   $\begin{bmatrix} ..\text{LS } \boxed{5} \begin{bmatrix} \text{L-MAX } \boxed{l_{max}} \\ \text{L-MIN } \boxed{l_{verb}} \end{bmatrix} \\ \text{SUBORD } \boxed{3} \\ \text{CONDS } \boxed{1} \end{bmatrix}$

**Underspecified Representations for Plural**
We argued that for an underspecified representation of plural NPs as regards the collective/distributive ambiguity, their meaning has to be represented by *potentially scope bearing* partial DRSs. This was achieved by stating the minimal label of the plural NP to be *weakly subordinated* to its maximal label in (6). Yet, in order to allow for an underspecified representation of the example given in (9), the lexical entry of the verb, stated in (2), has to be refined as indicated in (10).

(9) The lawyers hired a secretary.

$$(10) \begin{bmatrix} \text{CAT | H | SC } < \begin{bmatrix} \text{CASE } \text{nom} \\ \text{UDRS } \boxed{1} \end{bmatrix}, \begin{bmatrix} \text{CASE } \text{acc} \\ \text{UDRS } \boxed{2} \end{bmatrix} > \\ \text{UDRS} \begin{bmatrix} \text{LS } [\text{L-MIN } \boxed{l}] \\ \text{SUBORD } \{\} \\ \text{CONDS } \left\{ \begin{bmatrix} \text{LABEL } \boxed{l} \\ \text{REL } \text{hire} \\ \text{ARG1 } \text{dref\_res}(\boxed{1}, \text{Cond1}) \\ \text{ARG2 } \text{dref\_res}(\boxed{2}, \text{Cond2}) \end{bmatrix} \right\} \end{bmatrix} \end{bmatrix}$$

Note that as long as it is not determined whether a distributive or collective reading will be chosen for the plural NP, the discourse referent which occupies the corresponding argument place of the verb cannot be identified with the group referent introduced by the plural NP *the lawyers*. Instead, the mapping between NP meanings and the corresponding argument slots of the verb will be defined by a function *dref_res*, which returns the value of the appropriate discourse referent once a particular plural interpretation is chosen for (9).

But as long as the plural ambiguity is unresolved the function *dref_res* will be undefined. Thus, if context does not provide us with further, disambiguating information, (11) will be the final, underspecified representation for (9). Here, the function *dref_res* is undefined for the (underspecified) plural subject NP.

$$(11) \begin{bmatrix} \text{SUB } \{1_\top \geq \boxed{l_1}, 1_\top \geq \boxed{l_2}, \boxed{l_1} \geq \boxed{l_{12}}, \boxed{l_{12}} \geq \boxed{l_3}, \boxed{l_2} \geq \boxed{l_3}\} \\ \text{CONDS } \left\{ \begin{bmatrix} \text{LABEL } \boxed{l_1} \\ \text{REL } \text{lawyers} \\ \text{DREF } X \end{bmatrix}, \begin{bmatrix} \text{LABEL } \boxed{l_2} \\ \text{REL } \text{secr.} \\ \text{DREF } y \end{bmatrix}, \begin{bmatrix} \text{LABEL } \boxed{l_3} \\ \text{REL } \text{hire} \\ \text{ARG1 } \text{dref\_res}(\text{UDRS1}, \text{Cond1}) \\ \text{ARG2 } y \end{bmatrix} \right\} \end{bmatrix}$$

Note that the requirement for an underspecified representation of the discourse referent to fill the argument place of the verb cannot be implemented by use of a type hierarchy or similar devices which come to mind straightforwardly. For it is *not* appropriate for the issue of underspecified representations to compute the set of disjunctive readings, which would ensue automatically if we took such an approach. Instead, the function *dref_res* will be implemented by using delaying techniques. The conditions which determine the delayed evaluation of the function *dref_res* are defined in its second argument *Cond*. As long as the variable *Cond* is not instantiated, the evaluation of *dref_res* will be blocked, i.e. *delayed*.[13]

The three clauses of the function *dref_res* in (12) and (13) distinguish between *not scope bearing*, *scope bearing* and *potentially scope bearing elements*.

$$(12) \quad \text{dref\_res} \left( \begin{bmatrix} \text{LS } \begin{bmatrix} \text{L-MAX } \boxed{l_1} \\ \text{L-MIN } \boxed{l_{12}} \end{bmatrix} \\ \text{SUBORD}\{..\boxed{2} [\boxed{l_1} = \boxed{l_{12}}]..\} \\ \text{CONDS } \left\{ .. \begin{bmatrix} \text{LABEL } \boxed{l_1} \\ \text{DREF } \boxed{x} \end{bmatrix} .. \right\} \end{bmatrix}, \boxed{2} \right) := \boxed{x}$$

$$\text{dref\_res} \left( \begin{bmatrix} \text{LS } \begin{bmatrix} \text{L-MAX } \boxed{l_1} \\ \text{L-MIN } \boxed{l_{12}} \end{bmatrix} \\ \text{SO}\{..\boxed{2} [\boxed{l_1} > \boxed{l_{12}}], \boxed{l_1} > \boxed{l_{11}}..\} \\ \text{CONDS } \left\{ .. \begin{bmatrix} \text{LABEL } \boxed{l_{11}} \\ \text{DREF } \boxed{x} \end{bmatrix} .. \right\} \end{bmatrix}, \boxed{2} \right) := \boxed{x}$$

The first clause of (12), which takes as its first argument the UDRS value of a verb argument, as defined in (10), is only appropriate for non-quantificational singular NPs (4). The SUBORD value pertaining to the argument is constrained to contain a condition which identifies its minimal and maximal labels: $l_1 = l_{12}$. The second clause applies if the semantic structure of the argument contains a subordination restriction which characterizes the NP as *scope bearing*. This is the case for generalized quantifiers (3). The values of the minimal and maximal labels are characterized as non-identical by strong subordination: $l_1 > l_{12}$.

---

[13]In the CUF system (Doerre/Dorna 1993) delay statements are defined by the predicate *wait*. The delayed function can only be evaluated when all specified argument positions are instantiated. The delay statement for *dref_res* is *wait*(*dref_res*(*udrs*, *subord_info*)), where *subord_info* is the type of a member of SUBORD.

If a clause is applied successfully, by coindexation of the differentiating subordination restrictions with the second argument of *dref_res*, the latter gets properly instantiated and the function is relieved from its delayed status. It returns the discourse referent which in the argument's UDRS is associated with the maximal label for *not scope bearing* NPs, and with the label of the restrictor $l_{11}$ for *scope bearing* NPs. For plural NPs, which are represented as *potentially scope bearing* by a weak subordination constraint as shown in (6), the clauses in (12) will fail: the required subordination conditions will not be contained in the SUBORD value of the verb argument.[14] Underspecified as well as disambiguated plural NPs, characterized by a *weak* subordination constraint in the local UDRS, are captured by the third clause of *dref_res* in (13).

$$(13)\ \mathrm{dref\_res}\left(\begin{bmatrix}\mathrm{LS}\begin{bmatrix}\mathrm{L\text{-}MAX}\ \boxed{l_1}\\ \mathrm{L\text{-}MIN}\ \boxed{l_{12}}\end{bmatrix}\\ \mathrm{SUBORD}\{..\boxed{l_1}\geq\boxed{l_{12}}..\}\end{bmatrix},\mathrm{Cond}\right) := \_$$

In (13) the value of *dref_res* is undefined ($\top$) and the variable *Cond*, which is subject to the delay conditions on *dref_res*, is not instantiated by coindexation with a subordination restriction in the local SUBORD value. The function therefore is delayed, until further disambiguating constraints are available which resolve the plural ambiguity and determine the discourse referent to fill the argument slot of the verb. This is what we aimed at for the special concerns of plural underspecification.

If, however, a particular reading of a plural NP is determined by the lexical meaning of the verb, as it is the case for *gather*, an appropriate definition of *dref_res* in the lexical entry of the verb ensures the correct plural interpretation.

**Plural Disambiguation** In most cases, however, disambiguating information for the interpretation of plurals comes from various sources of semantic or pragmatic knowledge. Usually it is provided by subsequent discourse. We therefore define a mechanism for plural disambiguation which may apply at *any* stage of the derivation, to add disambiguating DRS conditions and subordination constraints to the underspecified representation whenever enough information is available to determine a particular plural interpretation. To this end we extend the Semantics Principle to include a function *pl_dis* (*plural disambiguation*), which applies to a phrase's UDRS value, to render a new value of the same type, which specifies a collective or distributive reading for a plural discourse referent contained in the underspecified representation. The individual clauses of *pl_dis* will have to state constraints for determining the respective plural readings, to be satisfied by the preceding context, represented in UDRS. Ideally, these constraints have access to inference modules, including semantic and pragmatic knowledge. We first state the function *pl_dis* for the different readings and then incorporate the function into the Semantics Principle.

If in clause (14) of *pl_dis* the constraints that determine a collective reading of the plural NP with label $l_1$ are satisfied, the relation of weak subordination between the minimal and maximal label of the plural NP is strenghtened to the identity relation. In the output value the restriction $l_1 = l_{12}$ gets unioned to the original SUBORD value. Note that the function *pl_dis* is fully monotonic in that its result is a UDRS which is obtained by only *adding* information to the input values SUBORD and CONDS by union.

Whenever disambiguation of a plural NP takes place, the function *dref_res* must be relieved from its delayed status in order to instantiate the corresponding argument slot of the verb. We will access the delayed goal *dref_res* by reference to the plural NP's maximal and minimal labels $l_1$ and $l_{12}$, instantiate its second argument by the identity constraint $l_1 = l_{12}$, and define its value by the DREF value **X** associated with $l_1$. The resulting UDRS for a collective interpretation of (9) is given in (15).

$$(14)\ \mathrm{pl\_dis}\left(\begin{bmatrix}\mathrm{LS}\ \boxed{3}\\ \mathrm{SUBORD}\ \boxed{2}\ \{..,\boxed{l_1}\geq\boxed{l_{12}},..\}\\ \mathrm{CONDS}\ \boxed{1}\ \left\{..,\begin{bmatrix}\mathrm{LABEL}\ \boxed{l_1}\\ \mathrm{DREF}\ \boxed{X}\end{bmatrix},..\right\}\end{bmatrix}\right) :=$$
$$\begin{bmatrix}\mathrm{LS}\ \boxed{3}\\ \mathrm{SUBORD}\ \boxed{2}\cup\{\boxed{4}\ [\boxed{l_1}=\boxed{l_{12}}]\}\\ \mathrm{CONDS}\ \boxed{1}\end{bmatrix}$$

Conditions: constraints for a collective reading (of X) &
$\exists$ delayed-goal: $\mathrm{dref\_res}\left(\begin{bmatrix}\mathrm{LS}\begin{bmatrix}\mathrm{L\text{-}MAX}\ \boxed{l_1}\\ \mathrm{L\text{-}MIN}\ \boxed{l_{12}}\end{bmatrix}\end{bmatrix},\boxed{4}\right)=\boxed{X}$

$$(15)\begin{bmatrix}\mathrm{SUBORD}\ \{1_\top\geq\boxed{l_1},1_\top\geq\boxed{l_2},\boxed{l_1}\geq\boxed{l_{12}},\boxed{l_1}=\boxed{l_{12}},\\ \boxed{l_{12}}\geq\boxed{l_3},\boxed{l_2}\geq\boxed{l_3}\ \}\\ \mathrm{CONDS}\ \left\{\begin{bmatrix}\mathrm{LABEL}\ \boxed{l_1}\\ \mathrm{REL}\ lawyers\\ \mathrm{DREF}\ \boxed{X}\end{bmatrix},\begin{bmatrix}\mathrm{LABEL}\ \boxed{l_2}\\ \mathrm{REL}\ secr.\\ \mathrm{DREF}\ \boxed{Y}\end{bmatrix},\begin{bmatrix}\mathrm{LABEL}\ \boxed{l_3}\\ \mathrm{REL}\ hire\\ \mathrm{ARG1}\ \boxed{X}\\ \mathrm{ARG2}\ \boxed{Y}\end{bmatrix}\right\}\end{bmatrix}$$

Disambiguation to a distributive reading is obtained in (16) by adding a quantificational distribution condition to the original value of CONDS. The restrictor $l_{11}$ introduces an individual discourse referent **x** together with the distribution condition $\mathbf{x}\in\mathbf{X}$ and the nuclear scope is identified by the minimal label $l_{12}$. Moreover, (strong) subordination of restrictor and scope is defined in SUBORD. Again, the delayed function *dref_res* is defined to return the discourse referent **x** which is to fill the argument slot of the verb and is un-delayed by instantiation of its second argument.

---

[14]This will be so even if – by the function *pl_dis* to be introduced below – further disambiguating constraints for, e.g., a collective or distributive reading are introduced at a later stage of the derivation: *dref_res* is defined on the UDRS value of a verb argument in the lexical entry of the verb. The value of this local UDRS, and with it the SUBORD attribute, remains unaffected by the introduction of additional subordination restrictions by clauses of the Semantics Principle.

$$(16) \quad pl\_dis \left( \begin{bmatrix} \text{LS } \boxed{3} \\ \text{SUBORD } \boxed{2} \; \{..,\boxed{l_1} \geq \boxed{l_{1_2}},..\} \\ \text{CONDS } \boxed{1} \; \{.., \begin{bmatrix} \text{LABEL } \boxed{l_1} \\ \text{DREF } \boxed{X} \end{bmatrix},..\} \end{bmatrix} \right) :=$$

$$\begin{bmatrix} \text{LS } \boxed{3} \\ \text{SUBORD } \boxed{2} \cup \{ \boxed{l_1} > \boxed{l_{1_1}}, \boxed{4} \; [\boxed{l_1} > \boxed{l_{1_2}}] \} \\ \text{CONDS } \boxed{1} \cup \left\{ \begin{bmatrix} \text{LABEL } \boxed{l_1} \\ \text{REL } \Rightarrow \\ \text{RES } \boxed{l_{1_1}} \\ \text{SCOPE } \boxed{l_{1_2}} \end{bmatrix}, \begin{bmatrix} \text{LABEL } \boxed{l_{1_1}} \\ \text{DREF } \boxed{x} \\ \text{REL } \in \\ \text{ARG1 } \boxed{X} \\ \text{ARG2 } \boxed{X} \end{bmatrix} \right\} \end{bmatrix}$$

Conditions:
constraints for a distributive reading (of X) &
$\exists$ delayed-goal: $dref\_res \left( \begin{bmatrix} \text{LS } \begin{bmatrix} \text{L-MAX } \boxed{l_1} \\ \text{L-MIN } \boxed{l_{1_2}} \end{bmatrix} \end{bmatrix}, \boxed{4} \right) = \boxed{x}$

We now complete the Semantics Principle by the Principle for Plural Disambiguation (VI). In (17), the function *pl_dis* applies in a coordination structure *coord-struc*, which recursively combines pairs of (sequences of) sentences and a sentence. The function *pl_dis* applies to the phrase's UDRS value, which is defined by application of the basic clauses (I) and (II) of UDRS construction. Depending on the context represented in UDRS, and supplemented by general semantic and/or pragmatic knowledge, *pl_dis* monotonically redefines the phrase's UDRS value if disambiguating constraints for a specific plural reading can be determined. If the constraints for plural disambiguation (14) and (16) are not satisfied, the trivial clause of *pl_dis* applies, which returns the UDRS value of its argument without modifications.

**Semantics Principle:** Clauses I, II and VI

$$(17) \quad \begin{bmatrix} ..\text{UDRS} \;\; pl\_dis \left( \begin{bmatrix} \text{SUBORD } \boxed{3} \cup \boxed{4} \\ \text{CONDS } \boxed{1} \cup \boxed{2} \end{bmatrix} \right) \\ coord-struc \end{bmatrix}$$

COORD-DTR          COORD-DTR

$\left[ ..\text{UDRS} \begin{bmatrix} \text{SUBORD } \boxed{4} \\ \text{CONDS } \boxed{2} \end{bmatrix} \right]$ $\left[ ..\text{UDRS} \begin{bmatrix} \text{SUBORD } \boxed{3} \\ \text{CONDS } \boxed{1} \end{bmatrix} \right]$

## 5 Conclusion and Perspectives

A constraint based semantic formalism for HPSG has been presented to replace the standard HPSG semantics. The new formalism comes closer to a principle based construction of semantic structure and, therefore, is more in the spirit of HPSG philosophy than its standard approach. Furthermore the new formalism overcomes a number of shortcomings of the standard approach in a natural way.

In particular, we presented an HPSG grammar for English that defines a syntax-semantics interface for the construction of U(nderspecified) D(iscourse) R(epresentation) S(tructure)s. The construction is guided by general principles, which clearly identify the interaction between the modules, i.e. the "interface" between syntax and semantics. In the fragment we defined underspecified representations for quantificational structures and plural NPs. The principles governing the interaction of syntax and semantics specify scoping relations for quantifiers and quantificational readings of plural NPs.

In addition to the syntax/semantics interface the Semantics Principle developed in this paper defines a clear interface to contextual and pragmatic knowledge. This interface allows reasoning modules to interact with semantics construction. The approach taken here can, therefore, be generalized to disambiguation problems other than the collective/distributive ambiguity as well as to anaphora resolution. A further issue to which the present account is directly related is incremental interpretation.